\documentclass[10pt,twocolumn,letterpaper]{article}

\usepackage{cvpr}
\usepackage{times}
\usepackage{epsfig}
\usepackage{graphicx}
\usepackage{amsmath}
\usepackage{amssymb}

\usepackage{subfigure}
\usepackage{amsthm}
\usepackage{wrapfig}
\usepackage{multirow}
\usepackage{capt-of}
\usepackage{booktabs}

\usepackage[breaklinks=true,bookmarks=false,colorlinks]{hyperref}

\cvprfinalcopy 

\def\cvprPaperID{5700} 

\ifcvprfinal\fi
\begin{document}

\title{Cross-Spectral Face Hallucination via Disentangling Independent Factors}

\author{
Boyan Duan$^{1}$\thanks{Equal Contribution} \quad
Chaoyou Fu$^{1,2*}$ \quad
Yi Li$^{1,2}$ \quad
Xingguang Song$^3$ \quad
Ran He$^{1,2}$\thanks{Corresponding Author} \\
$^1$ NLPR \& CEBSIT \& CRIPAC, CASIA \quad
$^2$ University of Chinese Academy of Sciences \\
$^3$ Central Media Technology Institute, Huawei Technology Co., Ltd. \\
{\tt\small 
dby96@163.com,
\{chaoyou.fu, rhe\}@nlpr.ia.ac.cn,
yi.li@cripac.ia.ac.cn,
songxingguang@huawei.com
}
}

\maketitle
\thispagestyle{empty}

\begin{abstract}
The cross-sensor gap is one of the challenges that have aroused much research interests in Heterogeneous Face Recognition (HFR). Although recent methods have attempted to fill the gap with deep generative networks, most of them suffer from the inevitable misalignment between different face modalities. Instead of imaging sensors, the misalignment primarily results from facial geometric variations that are independent of the spectrum. Rather than building a monolithic but complex structure, this paper proposes a Pose Aligned Cross-spectral Hallucination (PACH) approach to disentangle the independent factors and deal with them in individual stages. In the first stage, an Unsupervised Face Alignment (UFA) module is designed to align the facial shapes of the near-infrared (NIR) images with those of the visible (VIS) images in a generative way, where UV maps are effectively utilized as the shape guidance. Thus the task of the second stage becomes spectrum translation with aligned paired data. We develop a Texture Prior Synthesis (TPS) module to achieve complexion control and consequently generate more realistic VIS images than existing methods. Experiments on three challenging NIR-VIS datasets verify the effectiveness of our approach in producing visually appealing images and achieving state-of-the-art performance in HFR.
\end{abstract}

\section{Introduction}

\begin{figure}
  \centering
  \includegraphics[width=\linewidth]{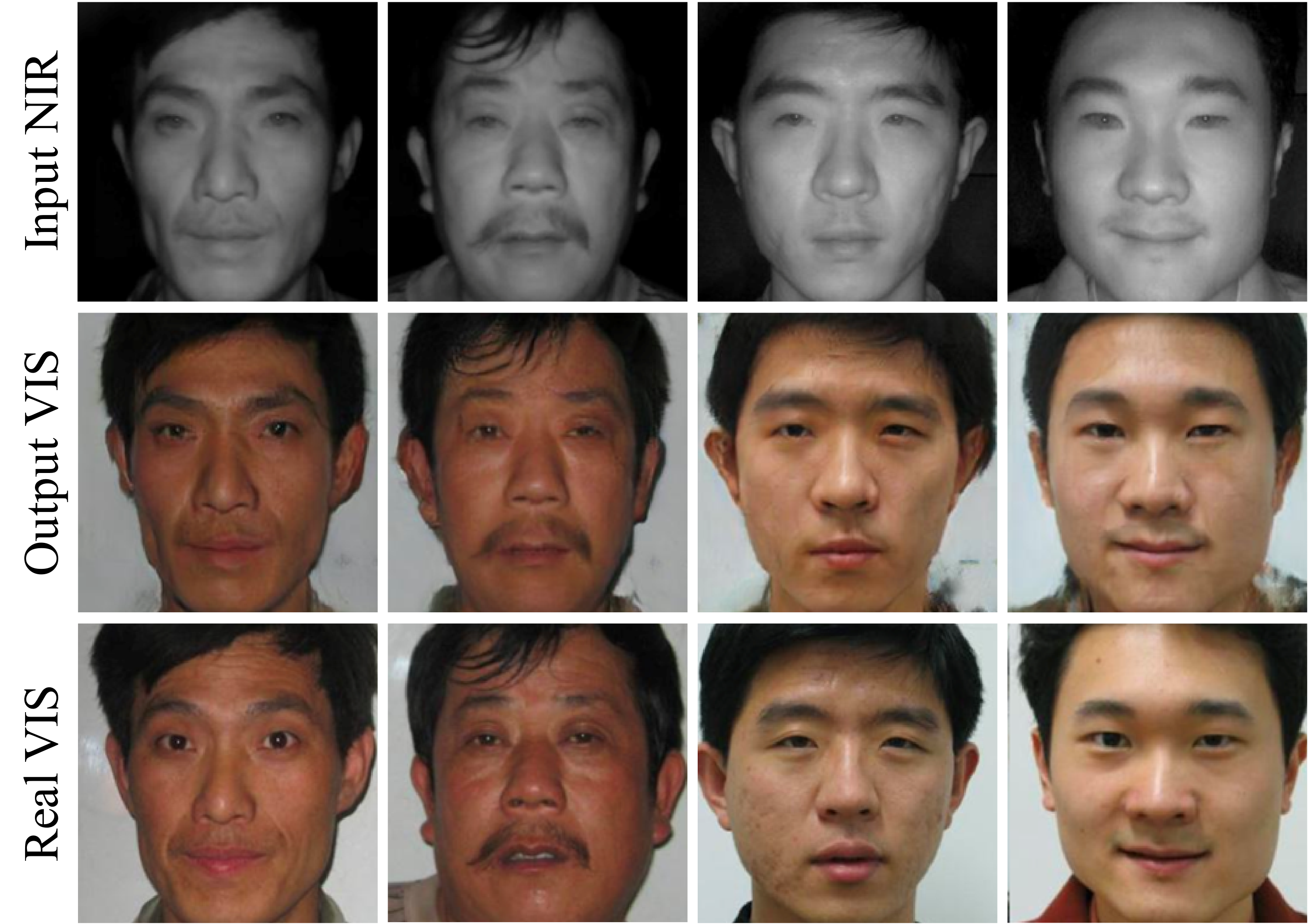}\\
  \caption{Synthesis results (the 2nd row, $256\times 256$ resolution) of PACH. There are distinct facial shape deviations between the NIR images (the 1st row) and the VIS images (the 3rd row). PACH disentangles the independent factors in cross-spectral hallucination and produces realistic VIS images from NIR inputs. }\label{pic1}
\end{figure}

In real world systems, there are multiple imaging sensors in cameras.
For example, near infrared (NIR) sensors work well in low lighting conditions and are widely used in night-vision devices and surveillance cameras.
Nevertheless, visible (VIS) images are much easier to capture, leading them to the most common type.
Different sensors result in face appearance variations, which imposes a great challenge to precisely match face images in different light spectra.
Face recognition with NIR images is an important task in computer vision \cite{liu2012heterogeneous}.
However, in most face recognition scenarios, the only available faces are VIS images.
There lack large-scale datasets with NIR faces for effective model learning, compared with the VIS face datasets.
Therefore, it is significant to effectively utilize both NIR and VIS images to boost HFR.
In past decades, many efforts have been paid to HFR.
These methods can be classified into three categories \cite{wu2019disentangled}.
The first category contrives to learn domain-invariant features of faces in different domains \cite{liao2009heterogeneous}.
The second category projects NIR and VIS images into a common subspace \cite{yi2007face}.
Face synthesis (or hallucination) has raised as another popular trend \cite{song2018adversarial}, especially in recent years.
It usually translates NIR images to the VIS ones while keeping the identity of faces, and then evaluates recognition models on the synthesized VIS images to reduce the domain gap.

\begin{figure*}
  \begin{center}
  \includegraphics[width=\linewidth]{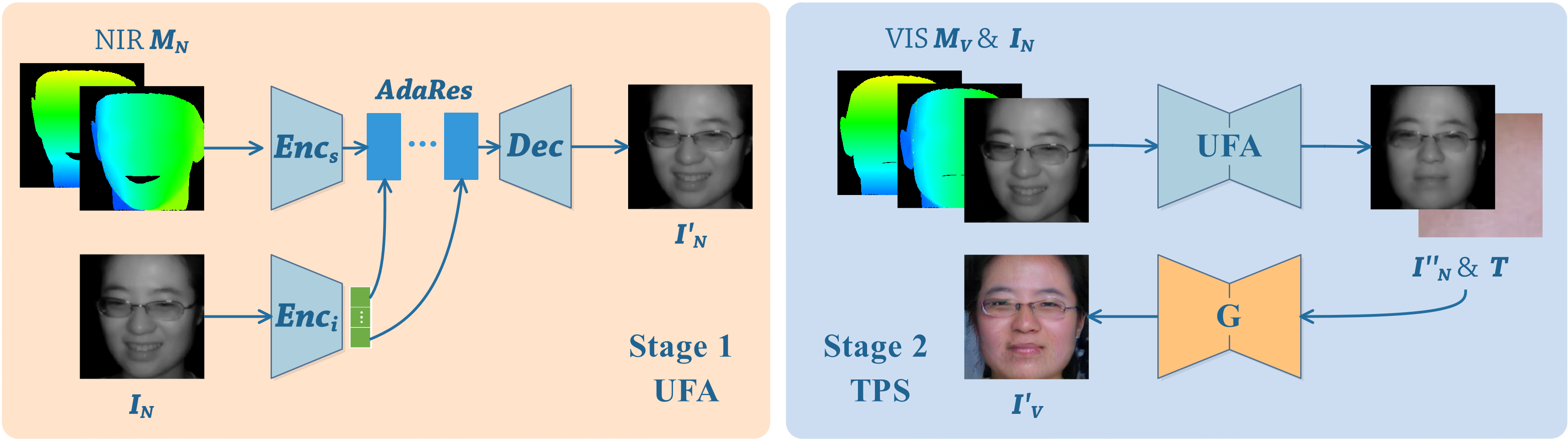}
  \end{center}
  \caption{The schematic diagram of PACH. There are two stages in our method, each having an individual duty. 
  The first stage (Unsupervised Face Alignment, UFA) learns to align the facial shape of $I_N$ to that of the paired $I_V$ with the guidance of the UV map.
  The second stage (Texture Prior Synthesis, TPS) transfers the aligned $I^{''}_N$ to a VIS one based on a texture prior $T$.}
  \label{pic2}
\end{figure*}

However, there are still challenges regarding the image synthesis based methods.
A major challenge comes from the misalignment.
The paired NIR and VIS images (coming from the same identity) in the training set are not exactly aligned.
The reason is that the NIR and the VIS images are usually captured in different scenarios, involving imaging distances or environments.
We exhibit some sample NIR-VIS pairs in Figure \ref{pic1}, along with our synthesized VIS results.
Nevertheless, most existing image synthesis methods require aligned paired data to train a decent model.
When confronting unaligned data (which is the more common case in reality), they tend to produce unsatisfying results.
In addition, the image resolution of the synthesized images is usually no more than $128 \times 128$.
Although \cite{yu2019ijcai} proposes to tackle the misalignment issue by learning attention from warped images to guide generation, their results share a similar complexion, which violates variations in reality and lacks realistic textures.
Moreover, their network is quite complex as well as with complicated data pre-processing.

In this paper, we propose a simple yet effective solution against the misalignment problem in cross-spectral face hallucination, namely Pose Aligned Cross-spectral Hallucination (PACH).
The schematic diagram is presented in Figure \ref{pic2}.
During the hallucination, procedures containing face alignment and spectrum translation are independent from each other.
Instead of dealing with the blended factors together, PACH disentangles them and settles each in an individual stage.
In the first stage, we design an Unsupervised Face Alignment (UFA) module to adjust the facial shape of the input NIR image.
UFA is trained following an unsupervised principle of reconstructing the input image.
Inspired by \cite{huang2018multimodal}, UFA could naturally separate the identity and the facial shape of an NIR image.
In the second stage, UFA has been trained well and stays unchanged.
The UV map of the input NIR image is replaced with that of the paired VIS image.
By this means, UFA synthesizes a new NIR image that is aligned with the paired VIS one.
The aligned paired data produced by UFA simplifies the task of cross-spectral hallucination.
To tackle the facial texture problem, we develop a Texture Prior Synthesis (TPS) module that is able to control complexions and produce realistic results.
We train our model on the CASIA NIR-VIS 2.0 dataset \cite{li2013casia}, and evaluate it on three datasets, including CASIA NIR-VIS 2.0, Oulu-CASIA NIR-VIS \cite{chen2009learning}, and BUAA-VisNir \cite{huang2012buaa}. 
Extensive experimental results show that our method generates high-quality images as well as promotes HFR performance.

In summary, our main contributions are as followings:

\begin{enumerate}
  \item This paper proposes a novel solution to deal with data misalignment in cross-spectral face hallucination, namely Pose Aligned Cross-spectral Hallucination (PACH). Since the facial shape and the spectrum are two independent factors, we suggest to disentangle the factors and settle them separately in different stages with relatively simpler networks.

  \item There are two stages in PACH, each focusing on a certain factor. In the first stage, we introduce an Unsupervised Face Alignment (UFA) module to adjust facial shape according to the guidance of the UV map, and thus produce aligned paired NIR-VIS data. The second stage contains a Texture Prior Synthesis (TPS) module that achieves complexion control and produces realistic VIS images for HFR.

  \item Extensive experiments on the CASIA NIR-VIS 2.0, the Oulu-CASIA NIR-VIS, and the BUAA-VisNir datasets show that our method achieves state-of-the-art performance in both visualization and recognition. The cross-dataset experiments demonstrate the generalization ability of our method.
\end{enumerate}

\section{Related Work}
Heterogeneous Face Recognition (HFR) has been widely studied in recent years. Existing methods could be classified into three categories: domain-invariant feature representation, common subspace learning, and image synthesis.

\textbf{Feature representation} methods try to learn face features that are robust and invariant in NIR and VIS domains.
Traditional methods are based on hand-crafted local features.
\cite{liao2009heterogeneous} applies Difference-of-Gaussian (DoG) filtering and Multi-scale Block Local Binary Patterns (MB-LBP) to get the feature representation.
\cite{galoogahi2012face} uses Local Radon Binary Pattern (LRBP) as the feature that is robust in two different modalities to tackle the task of Sketch-VIS recognition.
\cite{gong2017heterogeneous} encodes face images into a common encoding model, and uses a discriminant matching method to match images in different domains.

\textbf{Subspace learning} methods learn to project the NIR and the VIS images into a common subspace.
The projections of the same subject from two domains are similar in the subspace.
\cite{yi2007face} applies Canonical Correlation Analysis (CCA) learning in the Linear Discriminant Analysis (LDA) subspace. 
\cite{sharma2011bypassing} uses Partial Least Squares (PLS) to map heterogeneous faces from different modalities into a common subspace. 
\cite{huang2012regularized} proposes regularized discriminative spectral regression to match heterogeneous face images in a subspace.
\cite{kan2015multi} uses a Multi-view Discriminant Analysis (MvDA) approach to learn a discriminant common subspace.

\textbf{Image synthesis} methods aim to reduce the domain gap by a synthesis manner, e.g., translating NIR images to the VIS ones.
\cite{tang2003face} uses the image synthesis method to tackle the sketch-photo recognition problem.
\cite{juefei2015nir} learns a mapping function between the NIR and the VIS domains with a dictionary based approach.
In recent years, with the rise of deep learning, there are lots of works applying deep learning in the image synthesis process.
\cite{lezama2017not} uses a convolutional neural network to synthesize VIS images from NIR images in patches, and then applies a low-rank embedding to further improve the results.
Generative Adversarial Network (GAN) \cite{goodfellow2014generative} is also widely used in this field.
\cite{song2018adversarial} proposes to use a Cycle-GAN \cite{zhu2017unpaired} based framework for the face hallucination.
\cite{fu2019dual} proposes a dual generation method that generates massive paired NIR-VIS images from noise to reduce the domain gap of HFR.

\section{Method} \label{method}
\begin{figure}[t]
\begin{center}
\includegraphics[width=\linewidth]{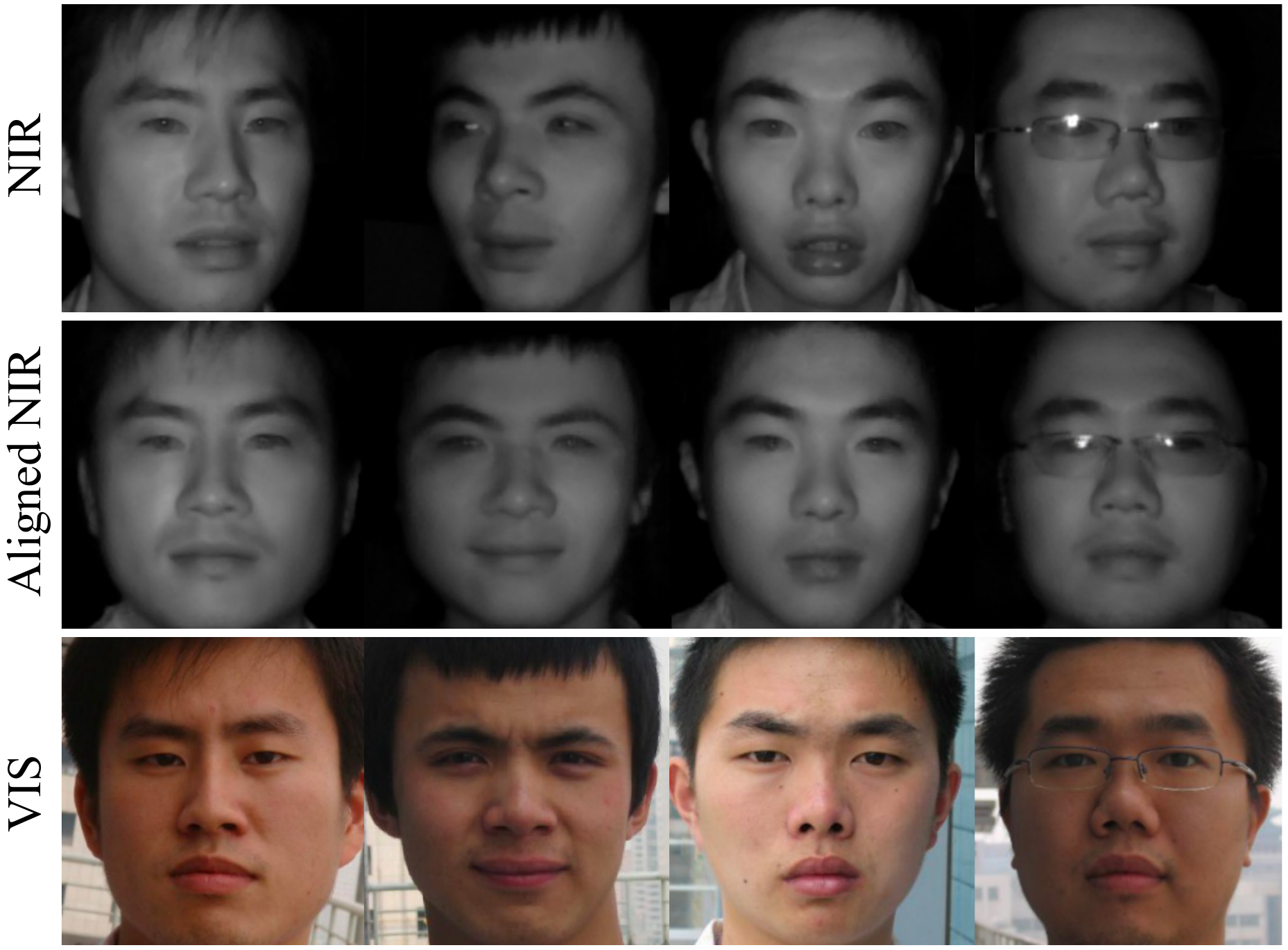}
\end{center}
\caption{Examples of face alignment on the CASIA NIR-VIS 2.0 dataset. There are large differences in the facial shapes between the paired NIR (the 1st row) and VIS (the 3rd row) images. The aligned NIR (the 2nd row) images have the same facial shapes as the VIS ones.}
\label{fig-1}
\end{figure}

The goal of our method is to translate an NIR image to the VIS one, which is expected to facilitate the performance of HFR.
However, on the one hand, the paired NIR and VIS images in the heterogeneous face datasets, such as CASIA NIR-VIS 2.0, are unaligned.
There are inevitable differences in the facial shapes between paired NIR and VIS images, as shown in Figure~\ref{fig-1}.
The misalignment of facial shapes makes it hard to synthesize satisfactory VIS images from the paired NIR ones.
On the other hand, the diverse complexions of the VIS images lead the NIR-VIS translation to be a `one to many' problem, i.e., one NIR complexion to multiple VIS complexions, bringing challenges to photo-realistic face synthesis.
In order to tackle the above problems, we explicitly divide the cross spectral face hallucination into two independent stages: an Unsupervised Face Alignment (UFA) stage and a Texture Prior Synthesis (TPS) stage.
The first stage is proposed to align the facial shapes of the NIR images with those of the paired VIS ones, as presented in Figure~\ref{fig-1}.
After that, we can obtain the aligned paired NIR and VIS images for pixel-wise supervised training.
The second stage adopts a texture prior to facilitate the realistic image synthesis.
In the following subsections, the details of the above two stages are described respectively.

\subsection{Unsupervised Face Alignment (UFA)}
Inspired by the recently proposed works \cite{huang2018multimodal,karras2019style,liu2019few,zakharov2019few} that employ Adaptive Instance Normalization (AdaIN) \cite{huang2017arbitrary} to control image styles, we propose an unsupervised face alignment method with AdaIN to disentangle facial shapes and identities.
The AdaIN is defined as:
\begin{equation} \label{eq-0}
    AdaIN(z, \gamma, \beta ) = \gamma \left ( \frac{z - u(z)}{\sigma(z)} \right ) + \beta. \\
\end{equation}
In the previous method \cite{huang2018multimodal}, $z$ means the feature of `content' images. $u(z)$ and $\sigma(z)$ denote the channel-wise mean and standard deviation of $z$, respectively. $\gamma$ and $\beta$ are the affine parameters learned by a network.
The image `style' can be switched by changing $\gamma$ and $\beta$.
In our method, we replace the `content' with the facial shape, and the `style' with the identity.

As shown in Figure~\ref{pic2}, the generator in UFA consists of a shape encoder $Enc_s$, an identity encoder $Enc_i$, several AdaIN residual blocks $AdaRes$, and a decoder $Dec$.
$Enc_i$ is used to extract the identity features, which are irrelevant with the facial shape, of the input NIR image $I_N$.
The affine parameters $\gamma$ and $\beta$ in Eq.~(\ref{eq-0}) are obtained by $Enc_i(I_N)$.
$Enc_s$ is a facial shape extractor. The input of $Enc_s$ is the UV map $M_N$ of $I_N$.
The facial shape of $I_N$, such as the pose and the expression, can be presented well by $M_N$, as shown in Figure~\ref{pic2}.
$AdaRes$, which denotes residual blocks with AdaIN, is used to disentangle the identity features $Enc_i(I_N)$ and the shape features $Enc_s(M_N)$.
$Dec$ decodes the disentangled features $AdaRes(Enc_i(I_N), Enc_s(M_N))$ to the image space, outputting the NIR image $I^{'}_N$.
The loss functions of this stage are introduced as below.

\subsubsection{Reconstruction Loss.}
We adopt an unsupervised manner to train the generator, which is reflected in the fact that we only reconstruct the input image without any other supervision.
The output image $I^{'}_N = Dec(AdaRes(Enc_i(I_N), Enc_s(M_N)))$ is required to keep consistent with the input image $I_N$, which is implemented by a pixel-wise L$1$ loss:
\begin{equation} \label{eq-1}
    \mathcal{L}_{\text{rec}} = \mathbb{E}_{I^{'}_N,I_N} [ || I^{'}_N - I_N ||_1]. \\
\end{equation}

\subsubsection{Identity Preserving Loss.}
Inspired by \cite{hu2018pose}, the reconstructed NIR image $I^{'}_N$ should be consistent with the ground truth $I_N$ not only at the image space, but also at the latent semantic feature space. Specifically, an identity preserving network $D_{ip}$, which is the LightCNN \cite{wu2018light} pre-trained on the MS-Celeb-$1$M dataset \cite{guo2016ms}, is introduced to extract the identity features of $I^{'}_N$ and $I_N$, respectively.
A L$2$ loss is imposed to constrain the feature distance between $D_{ip}(I^{'}_N)$ and $D_{ip}(I_N)$:
\begin{equation} \label{eq-2}
    \mathcal{L}_{\text{ip}} = \mathbb{E}_{I^{'}_{N},I_N} [ || D_{ip}(I^{'}_N) - D_{ip}(I_N) ||_2 ]. \\
\end{equation}

\subsubsection{Adversarial Loss.}
In order to improve the visual quality of the reconstructed NIR image $I^{'}_N$, we adopt a discriminator $D$ to perform adversarial learning \cite{goodfellow2014generative} with the generator, including $Enc_i$, $Enc_c$, $AdaRes$, and $Dec$:
\begin{equation} \label{eq-3}
\begin{aligned}
    \mathcal{L}_{\text{adv}} &= \mathbb{E}_{I_N} [ \log D(I_N)] + \mathbb{E}_{I^{'}_N} [ \log(1 - D(I^{'}_N)) ].
\end{aligned}
\end{equation}

\subsubsection{Overall Loss.}
The overall loss in the first stage is the weighted sum of the above reconstruction loss $\mathcal{L}_{\text{rec}}$, identity preserving loss $\mathcal{L}_{\text{ip}}$, and adversarial loss $\mathcal{L}_{\text{adv}}$:
\begin{equation} \label{eq-4}
    \mathcal{L}_{\text{UFA}} = \mathcal{L}_{\text{rec}} + \lambda_1 \mathcal{L}_{\text{ip}} + \lambda_2\mathcal{L}_{\text{adv}}. \\
\end{equation}
where $\lambda_1$ and $\lambda_2$ are trade-off parameters.
The generator, which contains $Enc_i$, $Enc_s$, $AdaRes$ and $Dec$, and the discriminator $D$ are trained alternatively to play a min-max game \cite{goodfellow2014generative}.

\subsection{Texture Prior Synthesis (TPS)}
After the training of UFA, we align the facial shape of the input NIR image $I_N$ with that of the target VIS image $I_V$, by changing the UV map. That is, replacing the UV map $M_N$ of $I_N$ with the UV map $M_V$ of $I_V$.
By this means, we obtain the aligned paired NIR-VIS training images $I^{''}_N$ = $Dec(AdaRes(Enc_i(I_N), Enc_s(M_V)))$ and $I_V$.
The corresponding process is presented in Figure~\ref{pic2}.
The examples of the aligned paired NIR and VIS images are shown in Figure~\ref{fig-1}.

By now we have aligned paired training data $I^{''}_N$ and $I_V$, but meet the other intractable challenge. 
The complexions of NIR images in the CASIA NIR-VIS 2.0 dataset are unified, while those of VIS images are diverse.
The diverse complexions lead cross-spectral hallucination to a `one to many' problem, which brings challenges to the traditional image to image translation methods \cite{isola2017image,zhu2017unpaired} that are usually applicable to `one to one' problems.
As shown in Figure~\ref{fig-3}, the synthesized images of previous translation methods tend to have an average complexion that is somewhat yellow. Obviously, the average complexion makes the synthesized images unrealistic, which may further decrease the recognition performance.

Different from previous image translation methods, we introduce a texture prior to facilitate cross-spectral hallucination.
Specifically, a texture prior $T$, which indicates the complexion information, is cropped from the target VIS image $I_V$. The texture prior is concatenated with the aligned NIR image $I^{''}_N$, and fed into the generator $G$.
By this way, $T$ provides a specific guidance for the translation, turning it into an easier ‘one to one’ task.
The corresponding losses in this stage are listed as follows.

\subsubsection{Pixel Loss.}
Benefitting from the aligned paired NIR-VIS images $I^{''}_N$ and $I_V$, we can train the translation network $G$ by the means of pixel-wise supervision.
The pixel loss, which defines the discrepancies between the synthesized $I^{'}_V = G(I^{''}_N, T)$ and the target $I_V$, is formulated as:
\begin{equation} \label{eq-5}
    \mathcal{L}_{\text{pix}} = \mathbb{E}_{I^{''}_N, T, I_V} [| G(I^{''}_N, T) - I_V |]. \\
\end{equation}

\subsubsection{Total Variation Regularization.}
In order to reduce the artifacts that are produced in the training process, a total variation regularization loss \cite{johnson2016perceptual} is imposed on the synthesized image: 
\begin{equation} \label{eq-6}
\begin{aligned}
    \mathcal{L}_{\text{tv}} =  \sum_{c=1}^{C} \sum_{w,h=1}^{W,H} & | G(I^{''}_N, T)_{w+1,h,c}  - G(I^{''}_N, T)_{w,h,c} | \\
    & + | G(I^{''}_N, T)_{w,h+1,c} - G(I^{''}_N, T)_{w,h,c} |. \\
\end{aligned}
\end{equation}
where $W$ and $H$ denote image width and image height, respectively.

In addition, we also adopt an identity preserving loss and an adversarial loss in this stage.
The two losses have the same form as Eq.~(\ref{eq-2}) and Eq.~(\ref{eq-3}) respectively, except for replacing $I_N/I^{'}_N$ with $I_V/I^{'}_V$.

\begin{figure}[t]
\begin{center}
\includegraphics[width=\linewidth]{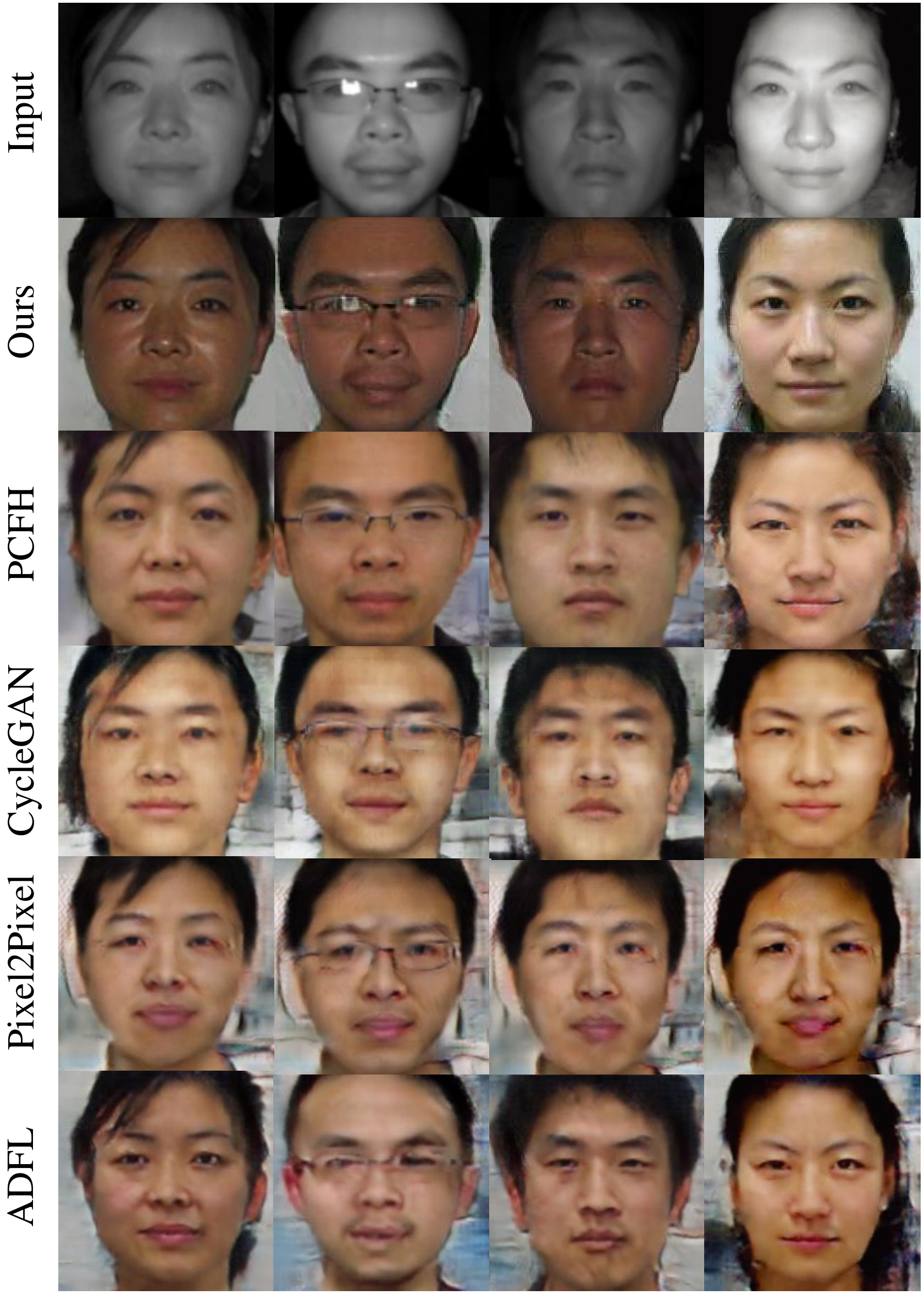}
\end{center}
\caption{The visualization comparisons with other state-of-the-art methods on the CASIA NIR-VIS 2.0 dataset. The results of the compared methods are obtained from \cite{yu2019ijcai}.
}
\label{fig-3}
\end{figure}

\subsubsection{Overall Loss.}
The overall loss in the second stage is the weighted sum of the above losses:
\begin{equation} \label{eq-7}
    \mathcal{L}_{\text{TPS}} = \mathcal{L}_{\text{pix}} + \alpha_1 \mathcal{L}_{\text{tv}} + \alpha_2 \mathcal{L}_{\text{ip}} + \alpha_3 \mathcal{L}_{\text{adv}}. \\
\end{equation}
where $\alpha_1$, $\alpha_2$, and $\alpha_3$ are the trade-off parameters.

\begin{table}[t]
    \begin{center}
    \resizebox{0.43\textwidth}{!}{
    \begin{tabular}{lccc}
        \toprule[0.9pt]
         Method & Rank-1 & VR@FAR=1\% & VR@FAR=0.1\% \\
        \midrule
        LightCNN~\cite{wu2018light} & 96.84 & 99.10 & 94.68\\
        Pixel2Pixel~\cite{isola2017image} & 22.13 & 39.22 & 14.45 \\
        CycleGAN~\cite{zhu2017unpaired} & 87.23 & 93.92 & 79.41 \\
        PCFH~\cite{yu2019ijcai} & 98.50 & 99.58 & 97.32 \\
        \midrule
        PACH & \textbf{99.00} & \textbf{99.61} & \textbf{98.51} \\
        \bottomrule[0.9pt]
    \end{tabular}}
    \end{center}
    \caption{Comparisons with other state-of-the-art methods on the $1$-fold of the CASIA NIR-VIS 2.0 dataset.}
    \label{table-1}
\end{table}

\begin{table}[t]
    \begin{center}
    \resizebox{0.47\textwidth}{!}{
    \begin{tabular}{lccc}
        \toprule[0.9pt]
        Method & Rank-1 & VR@FAR=1\% & VR@FAR=0.1\% \\
        \midrule
        VGG~\cite{parkhi2015deep} & 62.1 $\pm$ 1.88 & 71.0 $\pm$ 1.25 & 39.7 $\pm$ 2.85\\
        TRIVET~\cite{liu2016transferring} & 95.7 $\pm$ 0.52 & 98.1 $\pm$ 0.31 & 91.0 $\pm$ 1.26 \\
        LightCNN~\cite{wu2018light} & 96.7 $\pm$ 0.23 & 98.5 $\pm$ 0.64 & 94.8 $\pm$ 0.43 \\
        IDR~\cite{he2017learning} & 97.3 $\pm$ 0.43 & 98.9 $\pm$ 0.29 & 95.7 $\pm$ 0.73 \\
        ADFL~\cite{song2018adversarial} & 98.2 $\pm$ 0.34 & 99.1 $\pm$ 0.15 & 97.2 $\pm$ 0.48 \\
        PCFH~\cite{yu2019ijcai} & 98.8 $\pm$ 0.26 & 99.6 $\pm$ 0.08 & 97.7 $\pm$ 0.26 \\
        \midrule
        PACH & \textbf{98.9 $\pm$ 0.19} & \textbf{99.6 $\pm$ 0.10} & \textbf{98.3 $\pm$ 0.21} \\
        \bottomrule[0.9pt]
    \end{tabular}}
    \end{center}
    \caption{Comparisons with other state-of-the-art methods on the $10$-fold of the CASIA NIR-VIS $2.0$ dataset.}
    \label{table-2}
\end{table}

\section{Experiments}
In this section, we evaluate our proposed approach against state-of-the-art methods on three widely employed NIR-VIS face datasets, including the CASIA NIR-VIS 2.0 \cite{li2013casia}, the Oulu-CASIA NIR-VIS \cite{chen2009learning}, and the BUAA-VisNir \cite{huang2012buaa} datasets. We begin with introducing these three datasets as well as the training and the testing protocols. Then, experimental details are described. Finally, qualitative and quantitative experimental results are reported to demonstrate the effectiveness of our approach.

\subsection{Datasets and Protocols} \label{Datasets}

\textbf{The CASIA NIR-VIS 2.0} \cite{li2013casia} is a challenging NIR-VIS heterogeneous face dataset with largest number of images from $725$ subjects.
The number of VIS images for each subject ranges from $1$ to $22$, and the number of NIR images for each subject ranges from $5$ to $50$.
Face images in this dataset contain diverse variations, such as different expressions, poses, backgrounds, and lighting conditions.
The paired NIR and VIS images of each subject are not aligned, because of the differences in facial shapes.
We follow the protocol of \cite{wu2019disentangled} to split the training and the testing set, containing a total of $10$-fold experimental settings.
For each setting, $2,500$ VIS images and $6,100$ NIR images from about $360$ subjects are used as the training set.
The probe set consists of over $6,000$ NIR images from $358$ subjects.
The gallery set contains $358$ VIS images from the same subjects.
Note that, we also follow the generation protocol of \cite{yu2019ijcai}. That is, the qualitative and quantitative results are all obtained from the first fold.
The Rank-1 accuracy, verification rate (VR)@ false accept rate (FAR) = $1$\%, and VR@FAR = $0.1$\% are reported for comparisons.

\textbf{The Oulu-CASIA NIR-VIS} \cite{chen2009learning} is a popular heterogeneous face dataset that consists of $80$ identities with $6$ different expressions.
Among all the identities, $30$ identities are from CASIA and the remainder are from Oulu University.
Following the protocol of \cite{wu2019disentangled}, $20$ identities are selected as the training set, and another $20$ identities are selected as the testing set.
Each identity contains $48$ NIR images and $48$ VIS images.
For the testing set, all the NIR images are used as the probe and all the VIS images are used as the gallery.
Following \cite{yu2019ijcai}, we train our model on the CASIA NIR-VIS 2.0 dataset and test it on the Oulu-CASIA NIR-VIS dataset.
The Rank-1 accuracy, VR@FAR = $1$\%, and, VR@FAR = $0.1$\% are reported.

\begin{table}[t]
    \begin{center}
    \resizebox{0.43\textwidth}{!}{
    \begin{tabular}{lccc}
        \toprule[0.9pt]
        Method & Rank-1 & VR@FAR=1\% & VR@FAR=0.1\% \\
        \midrule
        KDSR~\cite{huang2012regularized} & 66.9 & 56.1 & 31.9\\
        TRIVET~\cite{liu2016transferring} & 92.2 & 67.9 & 33.6 \\
        IDR~\cite{he2017learning} & 94.3 & 73.4 & 46.2 \\
        ADFL~\cite{song2018adversarial} & 95.5 & 83.0 & 60.7 \\
        LightCNN~\cite{wu2018light} & 96.7 & 92.4 & 65.1 \\
        PCFH~\cite{yu2019ijcai} & \textbf{100} & 97.7 & 86.6 \\
        \midrule
        PACH & \textbf{100} & \textbf{97.9} & \textbf{88.2} \\
        \bottomrule[0.9pt]
   \end{tabular}}
   \end{center}
   \caption{Comparisons with other state-of-the-art methods on the Oulu-CASIA NIR-VIS dataset.}
   \label{table-3}
\end{table}

\begin{table}[t]
    \begin{center}
    \resizebox{0.43\textwidth}{!}{
    \begin{tabular}{lccc}
        \toprule[0.9pt]
        Method & Rank-1 & VR@FAR=1\% & VR@FAR=0.1\% \\
        \midrule
        KDSR~\cite{huang2012regularized} & 83.0 & 86.8 & 69.5\\
        TRIVET~\cite{liu2016transferring} & 93.9 & 93.0 & 80.9 \\
        IDR~\cite{he2017learning} & 94.3 & 93.4 & 84.7 \\
        ADFL~\cite{song2018adversarial} & 95.2 & 95.3 & 88.0 \\
        LightCNN~\cite{wu2018light} & 96.5 & 95.4 & 86.7 \\
        PCFH~\cite{yu2019ijcai} & 98.4 & 97.9 & 92.4 \\
        \midrule
        PACH & \textbf{98.6}  & \textbf{98.0} & \textbf{93.5} \\
        \bottomrule[0.9pt]
    \end{tabular}}
    \end{center}
    \caption{Comparisons with other state-of-the-art methods on the BUAA-VisNir dataset.}
    \label{table-4}
\end{table}

\textbf{The BUAA-VisNir} \cite{huang2012buaa} is a widely used heterogeneous face recognition dataset.
It has images from $150$ subjects with $9$ NIR images and $9$ VIS images per subject.
A total of $50$ subjects with $900$ images are chosen as the training set, and the remaining $100$ subjects with $1800$ images are the testing set.
According to \cite{yu2019ijcai}, we train our model on the first fold of the CASIA NIR-VIS $2.0$ dataset, and test it on the BUAA-VisNir dataset.
The Rank-1 accuracy, VR@FAR = $1$\% and VR@FAR = $0.1$\% are reported.

\subsection{Experimental Details}
All images in the heterogeneous face datasets are aligned to $144 \times 144$ and center cropped to $128 \times 128$.
Moreover, we also align and crop $256 \times 256$ resolution images on the CASIA NIR-VIS 2.0 dataset to explore high-resolution face synthesis.
In the first stage, we crop a $15 \times 15$ patch from the facial cheek of the VIS image, and then resize it to $128 \times 128$ as the texture prior.
The UV map is calculated based on \cite{zhu2016face}.
Adam is used as the optimizer with a fixed learning rate 2e-4. The batch size is set to $64$.
Both of the trade-off parameters $\lambda_1$ and $\lambda_2$ in Eq.~(\ref{eq-4}) are set to $1$.
The trade-off parameters $\alpha_1$, $\alpha_2$, and $\alpha_3$ in Eq.~(\ref{eq-7}) are set to 1e-4, $1$, and $1$, respectively.
The network architectures in UFA are based on \cite{huang2018multimodal}, and those of the generator and the discriminator in TPS are based on \cite{hu2018pose}. Refer to Figure~\ref{pic2}, the input and output of the networks are modified correspondingly.

\begin{figure}[t]
\begin{center}
\includegraphics[width=\linewidth]{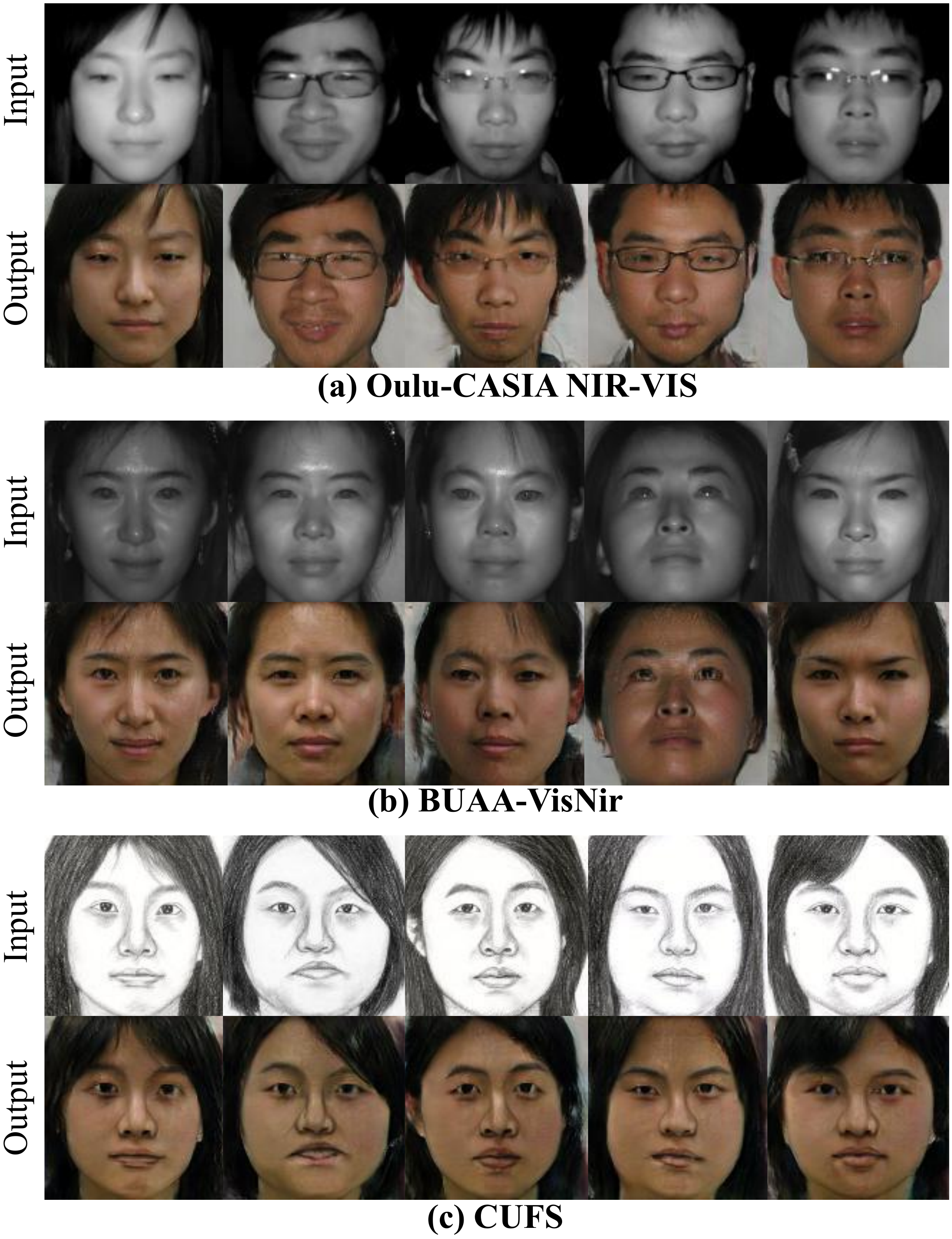}
\end{center}
\caption{The visualization results of cross-dataset experiments. The model is trained on the CASIA NIR-VIS 2.0 dataset. (a) The testing results on the Oulu-CASIA NIR-VIS dataset. (b) The testing results on the BUAA-VisNir dataset. (c) The testing results on the CUHK Face Sketch (CUFS) dataset \cite{wang2008face}.
}
\label{fig-5}
\end{figure}

\subsection{Comparisons}
\subsubsection{Results on the CASIA NIR-VIS 2.0 dataset.}
We compare the qualitative results of our method with those of other GAN-based methods, including Pixel2Pixel \cite{isola2017image}, CycleGAN \cite{zhu2017unpaired}, ADFL \cite{song2018adversarial}, and PCFH \cite{yu2019ijcai}, on the $1$-fold of the CASIA NIR-VIS 2.0 dataset. 
Among them, Pixel2Pixel and CycleGAN are well-known supervised and unsupervised image-to-image translation methods, respectively.
ADFL and PCFH are two state-of-the-art cross-spectral hallucination approaches. 
The visual comparisons are presented in Figure~\ref{fig-3}. 
All results of the compared methods are obtained from \cite{yu2019ijcai}.

For Pixel2Pixel and CycleGAN, there are distinct artifacts in the synthesized results. 
In addition, the facial shapes of the generated VIS images are not completely consistent with those of the input NIR images.
For example, the mouth shape of the second synthesized VIS image of CycleGAN is different from that of the input NIR image.
The facial size of the third synthesized VIS image of Pixel2Pixel is smaller than that of the input NIR image.
These phenomena may be caused by the unaligned paired training data.

ADFL is mainly based on CycleGAN, resulting in the similar visual problems as CycleGAN.
PCFH proposes a complex attention warping to alleviate the unaligned problem, and thus gets better results than Pixel2Pixel, CycleGAN, and ADFL.
However, there is still a huge gap between the synthesized images and the real ones, which is mainly reflected in the complexion.
The yellow complexion makes the results unrealistic.
It is obvious that our method outperforms all other methods. The synthesized VIS images not only maintain the facial shapes of the input NIR images, but also have more realistic textures.
We owe the consistency of facial shapes to the proposed face alignment in the first stage, and the realistic textures to the introduced texture prior in the second stage.

\begin{figure}[t]
\begin{center}
\includegraphics[width=\linewidth]{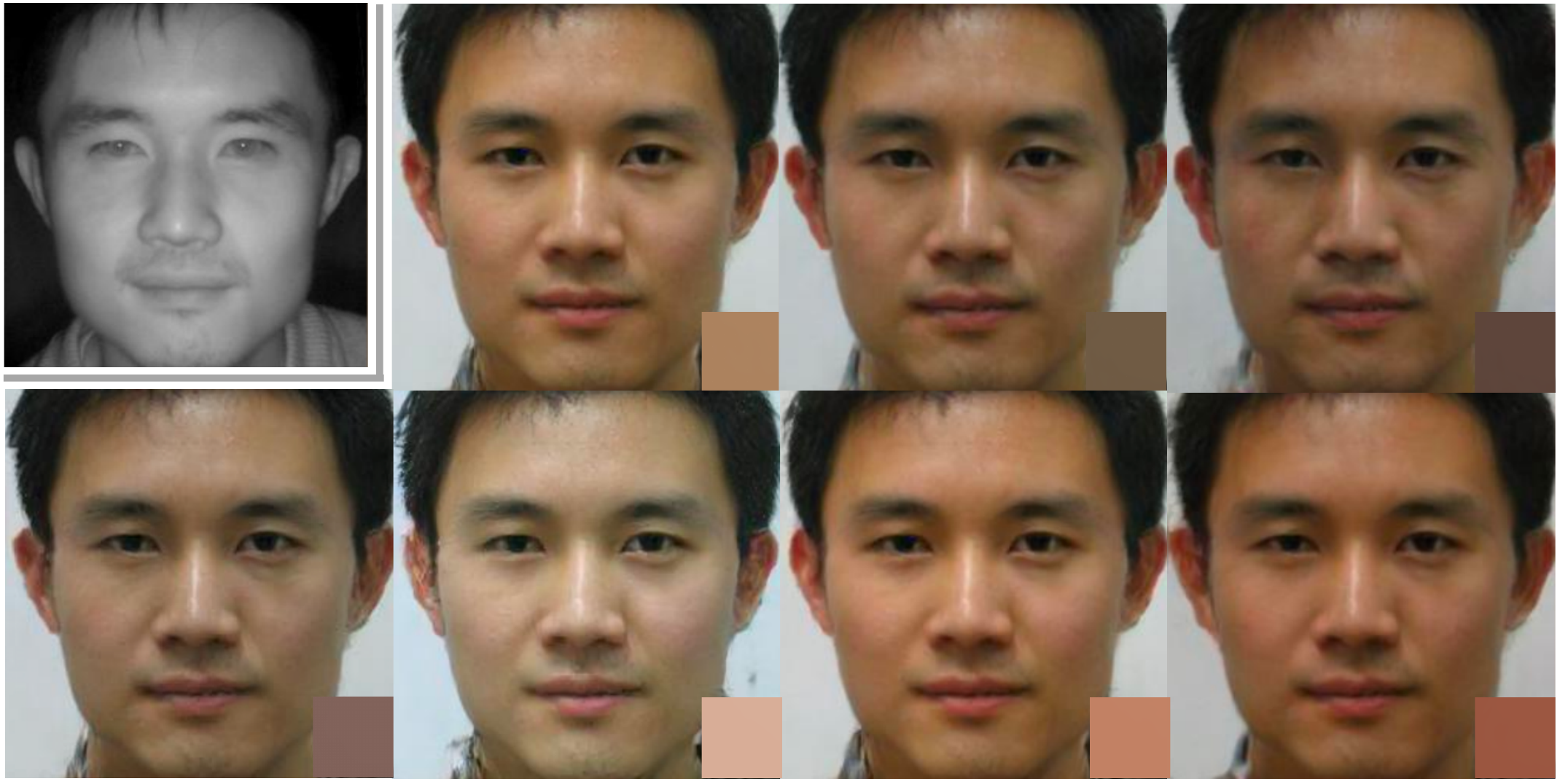}
\end{center}
\caption{The results of changing texture priors. The top left is the input NIR image. The rest VIS images are synthesized under different texture priors. For each synthesized VIS image, the lower right corner is the corresponding texture prior.
}
\label{fig-2}
\end{figure}

In Table~\ref{table-1}, we report the results of the quantitative comparison with Pixel2Piexl, CycleGAN, PCFH, and the baseline LightCNN on the 1-fold of the CASIA NIR-VIS 2.0 dataset.
We can see that our method performs better than the baseline LightCNN that evaluates on the original NIR images.
The Rank-1 accuracy, VR@FAR=1\%, and VR@FAR=0.1\% are improved by 2.16\%, 0.51\%, and 3.83\%, respectively.
The significant improvements over baseline prove that our method can really boost the recognition performance, by the way of translating NIR images to VIS ones.
On the contrary, compared with the baseline LightCNN, other GAN-based methods, i.e., Pixel2Pixel and CycleGAN, result in worse recognition performance.
The degradation may be caused by the poor quality of the synthesized images, as shown in Figure~\ref{fig-3}.

Furthermore, we also conduct experiments on more folds of the CASIA NIR-VIS 2.0 dataset, the results are tabulated in Table~\ref{table-2}.
Besides LightCNN, the compared methods contain VGG \cite{parkhi2015deep}, TRIVET \cite{liu2016transferring}, IDR \cite{he2017learning}, ADFL \cite{song2018adversarial}, and PCFH \cite{yu2019ijcai}.
Our method gets the best results on all recognition indicators. 
In particular, VR@FAR=0.1\% is improved from the state-of-the-art $97.7\%$ \cite{yu2019ijcai} to $98.3\%$.

\subsubsection{Results on the Oulu-CASIA NIR-VIS dataset.}
As stated in Section \ref{Datasets}, our model is trained on the 1-fold of the CAISA NIR-VIS 2.0 dataset, and tested on the Oulu-CASIA NIR-VIS dataset.
The qualitative cross-dataset experimental results are shown in Figure~\ref{fig-5} (a).
The input NIR images are randomly selected from Oulu-CASIA NIR-VIS.
We can observe our method still performs well in such a challenging cross-dataset case.

The results of the quantitative comparison with KDSR \cite{huang2012regularized}, TRIVET, IDR, ADFL, LighCNN, and PCFH are listed in Table~\ref{table-3}.
It is obvious that our method outperforms other methods by a large margin.
For instance, compared with the baseline LightCNN, VR@FAR=0.1\% is improved from 65.1\% to 88.2\%.
Compared with the state-of-the-art method PCFH, VR@FAR=0.1\% is improved by 1.6\%.
Since PCFH has got good performance in Rank-1 accuracy and VR@FAR=1\%, it is impressive to gain the improvements over PCFH.

\begin{table}[t]
    \begin{center}
    \resizebox{0.43\textwidth}{!}{
    \begin{tabular}{lccc}
        \toprule[0.9pt]
        Method & Rank-1 & VR@FAR=1\% & VR@FAR=0.1\% \\
        \midrule
        w/o UFA & 35.76 & 43.53 & 21.36  \\
        w/o TPS & 86.56 & 90.64 & 81.67 \\
        \midrule
        PACH & \textbf{99.00} & \textbf{99.61} & \textbf{98.51} \\
        \bottomrule[0.9pt]
    \end{tabular}}
    \end{center}
    \caption{Quantitative results of ablation study on the 1-fold of the CASIA NIR-VIS 2.0 dataset.}
    \label{table-5}
\end{table}

\subsubsection{Results on the BUAA-VisNir dataset.}
The cross-dataset experimental results on the BUAA-VisNir dataset are reported in Figure~\ref{fig-5} (b).
Our method obtains photo-realistic synthesized VIS images, although the model is trained on the CAISA NIR-VIS 2.0 dataset.

We further quantitatively compare our method with LightCNN, KDSR, TRIVET, IDR, ADFL, and PCFH.
The results of all the methods are shown in Table~\ref{table-4}. 
Compared with the baseline LightCNN, our method improves the Rank-1 accuracy, VR@FAR=1\%, and VR@FAR=0.1\% by 2.1\%, 2.6\%, and 6.8\%, respectively.
Moreover, compared with PCFH, our method gains 1.1\% on VA@FAR=0.1\%, revealing the importance of realistic textures.
The improvements on the Rank-1 and VR@FAR=1\% are marginal, because these indicators are already saturated.

\begin{figure}[t]
\begin{center}
\includegraphics[width=\linewidth]{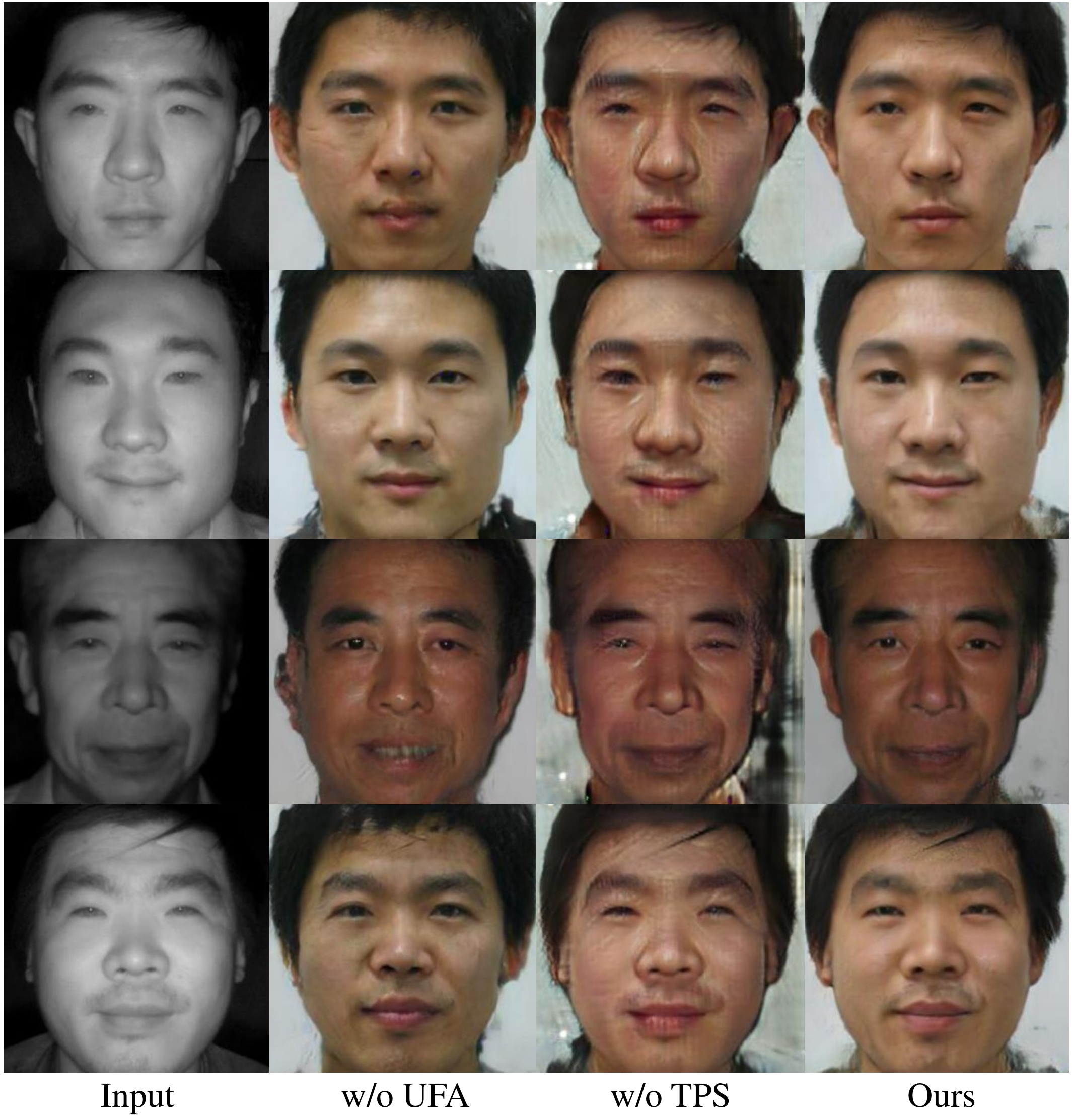}
\end{center}
\caption{The synthesis results of our method and its two variants on the CASIA NIR-VIS 2.0 dataset. The images in the first column are the input NIR images, and the remaining are the results without UFA, the results without TPS, and the results of our method, respectively.}
\label{fig-4}
\end{figure}

\subsection{Experimental Analyses}
We begin with studying the roles of our proposed UFA and TPS, reporting both qualitative and quantitative results for better comparisons.
Figure~\ref{fig-4} presents the visualization comparisons between our method and its two variants.
It is obvious that our method gets the best results.
Without UFA, the synthesized images are blurry, especially for the facial edges.
For example, the cheek of the first synthesized VIS image is not consistent with that of the input NIR image.
This may be caused by the unaligned paired data.
Without TPS, the synthesized images look unrealistic.
The diverse complexions of the VIS images in the CASIA NIR-VIS 2.0 dataset bring huge challenges for image translation.
Our texture prior provides a complexion simulation mechanism in the training process, facilitating to synthesize realistic facial textures.
Moreover, Figure~\ref{fig-2} shows the synthesized results under different textures priors.
The complexions of the synthesized results change with the texture priors, which demonstrates the controllability of the complexion.

Table~\ref{table-5} tabulates the quantitative recognition results of our method and its variants.
We can see that the recognition performance will highly decrease if any component is not used, suggesting that each component of our method is useful.
In particular, the recognition performance drops significantly when UFA is removed.
Concretely, the Rank-1 accuracy, VR@FAR=1\%, and VR@FAR=0.1\% decrease to 35.76\%, 43.53\%, and 21.36\%, when removing UFA.
The quantitative results in Table~\ref{table-5} further demonstrate the crucial role of our UFA and TPS for effective cross-spectral face hallucination.

In addition, we also make a parameter analysis, considering there are several trade-off parameters.
As stated in Section~\ref{method}, each loss of our method is reasonable, which is also confirmed by our experiments.
Specifically, when $\lambda_1$, $\lambda_2$, $\alpha_1$, $\alpha_2$, and $\alpha_3$ are set to $0$ respectively, the rank-1 accuracy on the CASIA NIR-VIS 2.0 dataset correspondingly decreases 14\%, 0.6\%, 0.5\%, 11\%, and 3\%.
Meanwhile, our method is not sensitive to these trade-off parameters in a large range.
For the most influential identity preserving loss, the rank-1 accuracy only changes 0.8\% when $\lambda_1$ is set from $1$ to $10$.

Given that our method performs well on the cross-dataset experiments, we further test our method on a sketch dataset CUHK Face Sketch (CUFS) \cite{wang2008face}. As shown in Figure~~\ref{fig-5} (c), although the model is only trained on the CASIA NIR-VIS 2.0 dataset, we observe satisfactory results on such a sketch dataset. The synthesized details, including the hair and the facial textures, are photo-realistic, which proves the generalization ability of our method. We will continue to explore more applications in our future work.

\section{Conclusion}
To tackle the misalignment problem in cross-spectral face hallucination, this paper has proposed to disentangle the facial shape and the spectrum information, and settle them in individual stages.
The first stage focuses on the facial shape. We design an Unsupervised Face Alignment (UFA) module to align the facial shape of an NIR image with that of the paired VIS one. 
Then we use the acquired aligned paired data to train a generator that translates the NIR image to VIS image.
The second stage is in charge of the cross-spectral translation.
To improve the reality of the synthesized results, we develop a Texture Prior Synthesis (TPS) module and produce VIS images with different complexion cases, which has been proved to facilitate the performance of cross-spectral translation.
We conduct extensive experiments on three challenging NIR-VIS datasets and achieve state-of-the-art results in visual effects and quantitative comparisons.

\section*{Acknowledgments}
This work is funded by Beijing Natural Science Foundation (Grants No. JQ18017).

{\small
\bibliographystyle{ieee_fullname}
\bibliography{references}
}

\end{document}